\documentclass[journal]{IEEEtran}
\usepackage{times}
\usepackage{epsfig}
\usepackage{graphicx}
\usepackage{amsmath}
\usepackage{amssymb}
\usepackage{epstopdf}
\usepackage{bm}
\usepackage{tabularx}
\usepackage{multirow}
\usepackage{subfigure}
\usepackage{threeparttable}

\begin{document}
\title{From Anchor Generation to Distribution Alignment: Learning a Discriminative Embedding Space for Zero-Shot Recognition}

\author{Fuzhen Li,~\IEEEmembership{,}
        Zhenfeng Zhu,~\IEEEmembership{,}
        Xingxing Zhang,~\IEEEmembership{,}
        Jian Cheng,~\IEEEmembership{,}
        and~Yao Zhao,~\IEEEmembership{}
\thanks{F. Zhu, F. Li , X. Zhang and Y. Zhao are with the Institute of Information Science, Department of Computer and Information Technology, Beijing Jiaotong University, Beijing, 100044, China, and also with the Beijing Key Laboratory of Advanced Information Science and Network Technology, Beijing, 100044, China (e-mail: zhfzhu@bjtu.edu.cn; lifuzhen@bjtu.edu.cn; zhangxing@bjtu.edu.cn; yzhao@bjtu.edu.cn).}
\thanks{J. Chen is with the National Laboratory of Pattern Recognition, Institute of Automation, Chinese Academy of Sciences (CAS), 100190, China (e-mail: jcheng@nlpr.ia.ac.cn).}
\thanks{}}

%
%

\markboth{Journal of \LaTeX\ Class Files,~Vol.~14, No.~8, August~2015}%
{Shell \MakeLowercase{\textit{et al.}}: Bare Demo of IEEEtran.cls for IEEE Journals}
%



\maketitle

\begin{abstract}
In zero-shot learning (ZSL), the samples to be classified are usually projected into side information templates such as attributes. However, the irregular distribution of templates makes classification results confused. To alleviate this issue, we propose a novel framework called Discriminative Anchor Generation and Distribution Alignment Model (DAGDA). Firstly, in order to rectify the distribution of original templates, a diffusion based graph convolutional network, which can explicitly model the interaction between class and side information, is proposed to produce discriminative anchors. Secondly, to further align the samples with the corresponding anchors in anchor space, which aims to refine the distribution in a fine-grained manner, we introduce a semantic relation regularization in anchor space. Following the way of inductive learning, our approach outperforms some existing state-of-the-art methods, on several benchmark datasets, for both conventional as well as generalized ZSL setting. Meanwhile, the ablation experiments strongly demonstrate the effectiveness of each component.
\end{abstract}


%
\IEEEpeerreviewmaketitle

\section{Introduction}
With the progress of deep convolutional neural network, significant success has been achieved in object recognition \cite{DBLP:conf/icml/DonahueJVHZTD14} \cite{DBLP:conf/cvpr/HeZRS16} \cite{DBLP:journals/corr/SimonyanZ14a} \cite{DBLP:conf/nips/KrizhevskySH12}. Although the performance has reached super-human level, the large-scale labeled samples are required for training supervised learning models. This dependence of deep learning models on labeled samples means that a lot of cost is invested in collecting and labeling data. To make matters worse, the frequencies of observed objects follow a long-tailed distribution \cite{DBLP:conf/cvpr/ZhuAR14}. For example, some categories such as endangered animals are difficult to collect. Therefore, in practice, it is unavoidable to recognize objects without any labeled training samples. Different from deep learning models, human can recognize unseen objects through transferring knowledge from relevant seen categories \cite{DBLP:conf/cvpr/FuS16}. Motivated by human learning mechanism, one-shot and zero-shot learning have attracted wide attention in recent years.

\begin{figure}[t]
\centering
\includegraphics[height=4.5cm,width=7.2cm]{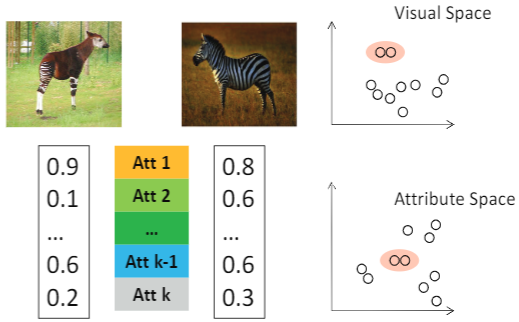}
\caption{An illustration of the knowledge transfer process in ZSL. The `Att' refers to attribute. If the zebra is similar to okapi in visual space, the similarity will be preserved in attribute space with only learning the projection from okapi to its attributes.}
\label{f1}
\end{figure}

Zero-Shot Learning (ZSL) \cite{DBLP:conf/cvpr/AkataPHS13} \cite{DBLP:conf/cvpr/AkataRWLS15} \cite{DBLP:conf/cvpr/FarhadiEHF09} \cite{DBLP:journals/corr/XianLSA17} is a popular learning paradigm to alleviate the above problems. In contrast to conventional supervised learning methods which require the emergence of test categories in the stage of training, ZSL learns from seen (source) categories to recognize unseen (target) categories as shown in Figure \ref{f1}. In the stage of training, existing ZSL approaches can be divided into two genres: inductive ZSL \cite{DBLP:conf/cvpr/AkataRWLS15} \cite{DBLP:conf/cvpr/ChangpinyoCGS16} \cite{DBLP:conf/nips/FromeCSBDRM13}, where only labeled samples are utilized for training, and transductive ZSL \cite{DBLP:conf/cvpr/SongSYLS18} \cite{DBLP:conf/nips/ZhaoDGLXW18} \cite{DBLP:journals/pami/FuHXG15}, where both labeled and unlabeled samples are utilized for training. In this paper, our method follows the inductive ZSL settting. In the stage of test, there are two evaluation criterions according to the scale of search space: 1) The conventional ZSL setting \cite{DBLP:conf/cvpr/SongSYLS18} refers to search in unseen categories; 2) The generalized ZSL setting \cite{DBLP:conf/eccv/ChaoCGS16} refers to search in both seen and unseen categories. In general, it is important to evaluate the performance of ZSL models under the conventional ZSL setting. However, in practical situations, we cannot ensure whether the test samples are from unseen classes. Thus, the generalized ZSL setting is usually a more convincing evaluation criterion in the real world.

In general, a shared semantic space is regarded as the bridge between seen and unseen categories, which is usually defined by attributes \cite{DBLP:conf/cvpr/AkataPHS13} \cite{DBLP:conf/cvpr/FarhadiEHF09}, word2vec \cite{DBLP:journals/corr/abs-1301-3781} and WordNet \cite{DBLP:journals/cacm/Miller95}. In the shared semantic space, the categories of unseen samples can be retrieved through the nearest neighbor search. However, the irregular distribution of templates makes classification results confused. To alleviate this problem, the existing approaches mainly try to learn a more discriminative shared embedding space \cite{DBLP:conf/pkdd/ShigetoSHSM15} or generate some unseen samples in the visual space \cite{DBLP:conf/cvpr/XianLSA18}.

In fact, ZSL can be transformed to link prediction problem on class-attribute bipartite graph. Graph Convolutional Network (GCN) \cite{DBLP:journals/corr/KipfW16}, which extends the conventional convolutional operation to graph, can effectively model the interaction between samples. By utilizing structure to aggregate sample information, GCN learns a more discriminative representation to support the downstream tasks such as node classification and link prediction. Furthermore, GCN can work well on generating discriminative templates.

\begin{figure*}[ht]
\centering
\includegraphics[height=7.5cm,width=15.5cm]{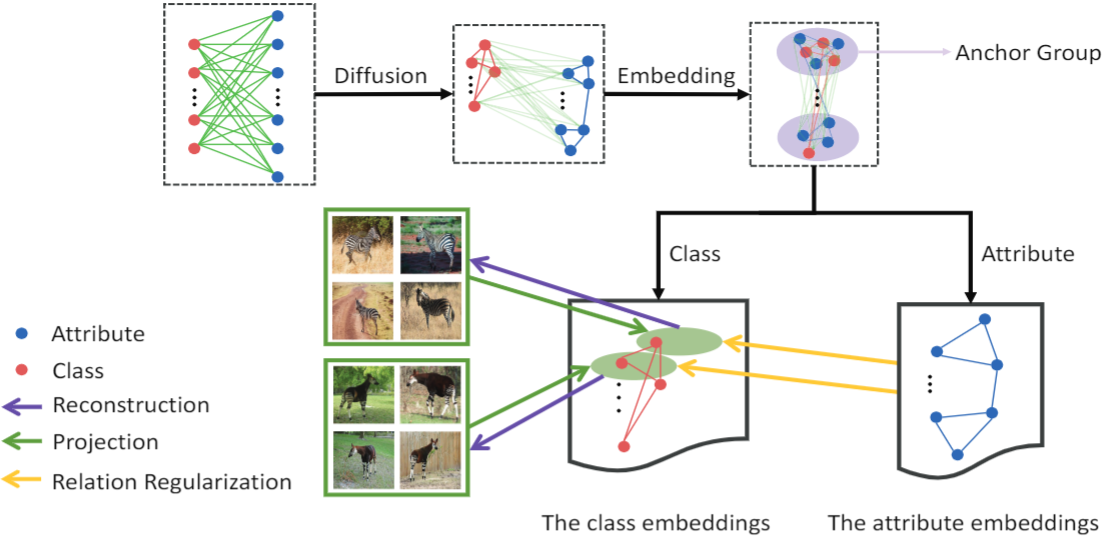}
\caption{A flowchart of the proposed DAGDA. The red points and the blue points separately refer to class and attribute. In the anchor generation stage, the discriminative anchor space can be discovered through a diffusion graph convolutional network.  With the learned anchors, the distribution alignment component projects samples from the feature space to the anchor space.}
\label{f2}
\end{figure*}

Based on aforementioned, we proposed a novel inductive ZSL framework in this paper. We dub the proposed ZSL framework as Discriminative Anchor Generation and Distribution Alignment Model (DAGDA), as it consists of anchor generation component and distribution alignment component as shown in Figure \ref{f2}. In the anchor generation component, we regard the classes and attributes as nodes in bipartite graph, and the edge denotes the interaction between class and attribute. We represent each node as a low-dimensional vector by GCN, which is the new template called anchor. Then the anchors are regarded as input of the distribution alignment component to classify the samples. Meanwhile, this component makes the distribution of samples in feature space consistent with that in anchor space.  

In summary, our contributions are:

$\bullet$ A novel inductive ZSL framework is proposed to achieve excellent performance in zero-shot recognition by generating discriminative anchors and aligning the sample distribution between the feature space and the anchor space. 

$\bullet$ A truncated diffusion graph convolutional network is proposed to effectively model the interaction between class and attribute and produce a discriminative anchor space. In addition, a relation regularization is employed to refine the distribution of samples.

$\bullet$ Theoretical derivation details how the truncated diffusion graph convolutional network works. Substantial experiments are carried out to demonstrate the effectiveness of our model and each component in it.

\section{Related Work}

\textbf{Zero-Shot Learning}\space\space Due to the absence of target classes in the training stage, the side information is a vital bridge that associates source and target classes. Thus, the definition of semantic space significantly effect the performance of the downstream ZSL methods. The side information usually includes attributes \cite{DBLP:conf/cvpr/AkataPHS13} \cite{DBLP:conf/cvpr/FarhadiEHF09} \cite{DBLP:conf/cvpr/LampertNH09}, word vector \cite{DBLP:conf/nips/FromeCSBDRM13} \cite{DBLP:journals/cacm/Miller95} and text description \cite{DBLP:conf/cvpr/ReedALS16} \cite{DBLP:conf/cvpr/ZhangXG17}. The attribute information is the most popular one which is certainly effective for ZSL \cite{DBLP:conf/cvpr/AkataRWLS15} \cite{DBLP:conf/cvpr/MorgadoV17}. According to the way how to associate sample features with attributes, the previous methods can be divided into three genres: 1) projecting sample features into the attribute space \cite{DBLP:conf/cvpr/AkataRWLS15} \cite{DBLP:conf/nips/FromeCSBDRM13} \cite{DBLP:conf/cvpr/ReedALS16}, 2) projecting attributes into the feature space \cite{DBLP:conf/iccv/KodirovXFG15} \cite{DBLP:conf/pkdd/ShigetoSHSM15} \cite{DBLP:conf/cvpr/ZhangXG17} and 3) projecting features and attributes into a shared latent space \cite{DBLP:conf/cvpr/ChangpinyoCGS16} \cite{DBLP:conf/ijcai/Lu16} \cite{DBLP:conf/cvpr/ZhangS16}. Once the projection is established, the top-1 classification result can be retrieve through the nearest neighbor searching in attribute space. However, since the distribution of attributes in attribute space is usually irregular ,which means it is difficult to carry out effective nearest neighbor searching. Our method aims to alleviate above problem through learning a discriminative shared space, which following the third genre.

\textbf{Inductive ZSL} \space\space In inductive ZSL, only the seen samples are available in the training stage. In Semantic Auto-encoder (SAE) \cite{DBLP:conf/cvpr/KodirovXG17}, the encoder projects the visual features into attribute space while the decoder is able to reconstruct the visual features, which can learn an efficient and robust projection function. F-CLSWGAN \cite{DBLP:conf/cvpr/XianLSA18} synthesizes sample features conditioned on attribute information and offers a shortcut directly from attribute to class-conditional feature distribution. \cite{DBLP:journals/prl/LongXSLXY18} learns the discriminative representations by utilizing Central-loss based Network (CLN) and maps class embeddings to representation space using Kernelized Ridge Regression (KRR). Preserving Semantic Relations (PSR) \cite{DBLP:conf/cvpr/AnnadaniB18} proposes to utilize the structure of the space spanned by the attributes. In this paper, we sufficiently utilize the interactions between classes and attributes to learn a discriminative anchor space.

\textbf{Graph Convolutional Network} \space\space GCN \cite{DBLP:journals/corr/KipfW16} is a landmark work extending convolution operation from image to graph, which can effectively aggregate neighborhood information. In \cite{DBLP:journals/corr/KipfW16}, the information aggregation occurs in each layer, the discriminative node representations can be produced during the layer-wise forward propagation. Although GCN is proposed for semi-supervised learning, it can be naturally extended to unsupervised learning manner. However, the scale of local information aggregation in \cite{DBLP:journals/corr/KipfW16} is small in each layer, which is insufficient to reveal the complex interactions between classes and attributes. In this paper, we develop a truncated diffusion graph convolution to capture the local structure more carefully. 

\begin{figure*}[ht]
\centering
\includegraphics[height=9cm,width=15cm]{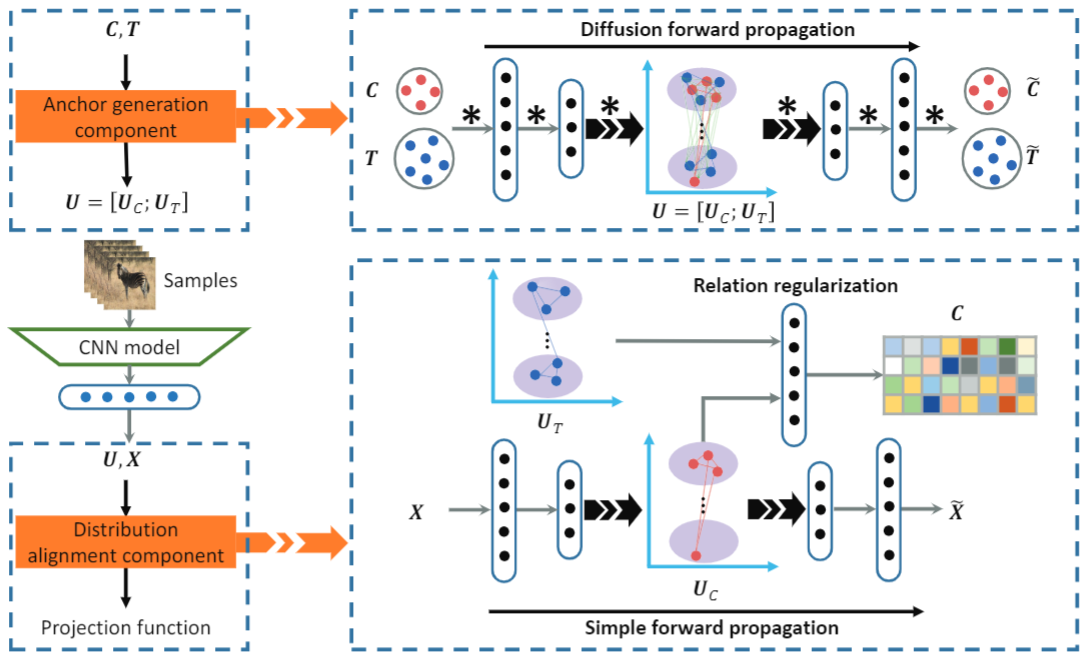}
\caption{An explicit depiction of the proposed framework. $*$ denotes graph convolution operation. In the anchor generation stage, the discriminative anchors $\bm{U}=[\bm{U}_C;\bm{U}_T]$ can be generated through the diffusion forward propagation conducted auto-encoder with the inputs $\bm{C}$ and $\bm{T}$. In the distribution alignment stage, a simple auto-encoder projects the $\bm{X}$ into the $\bm{U}$ space, while preserving the relations between sample features and anchors through a parameterized relation regularization network.}
\label{f3}
\end{figure*}
\section{Our Model}

\subsection{Notation and Problem Formulaiton}
Given a source dataset $\mathcal{D}^s= \{(\bm{x}_i^s, \bm{y}_i^s)\}_{i=1}^{N^s}$, each one $\bm{x}_i^s$ of $N^s$ images is corresponding to a label $\bm{y}_i^s$, $\bm{y}_i^s \in \mathcal{Y}^s = \{\bm{y}_i\}_{i=1}^{|s|}$, where $|s|$ is the number of the seen classes and $y_i$ is a one-hot encoding row vector. $\bm{X}^s = \theta(\bm{x}^s)$ denotes the visual features, and $\bm{Y}^s = \phi(\bm{y}^s)$ denotes the attributes bound to $\bm{X}^s$. Similar to the source dataset, there are $N^u$ unseen images in the target dataset $\mathcal{D}^u= \{(\bm{x}_i^u, \bm{y}_i^u)\}_{i=1}^{N^u}$, $\bm{y}_i^u \in \mathcal{Y}^u = \{\bm{y}_i\}_{i=|s|+1}^{|s|+|u|}$, where $|u|$ is the number of the unseen class. It is clear that $\mathcal{Y}^s\cap\mathcal{Y}^u=\emptyset$ and $\mathcal{Y}=\mathcal{Y}^s\cup\mathcal{Y}^u$. $\bm{X}^u = \theta(\bm{x}^u)$ denotes the visual features, and $\bm{Y}^u = \phi(\bm{y}^u)$ denotes the attributes bound to $\bm{X}^u$. $\bm{C} = [\phi(\bm{y}_1);\cdots;\phi(\bm{y}_{|s|});\phi(\bm{y}_{|s|+1});\cdots;\phi(\bm{y}_{|s|+|u|})]\in\mathcal{R}^{d_C\times d_T}$, where $d_C$ and $d_T$ separately refer to the number of classes and the number of attributes, denotes the class-attribute matrix. Naturally, $\bm{T}=\bm{C}^{\mathrm{T}}\in\mathcal{R}^{d_T\times d_C}$ is the attribute-class matrix. Generally, ZSL aims at minimizing the follow objective function:
\begin{equation}
\centering
Q = \frac{1}{N^s}\sum_{i=1}^{N^s}\|\phi(\bm{y}_i^s)-\phi(f(\bm{x}_i^s,\bm{W}))\|^2 + \lambda\Omega(\bm{W}),
\label{e1}
\end{equation}
where $f(\bm{x}_i^s,\bm{W})$ is a parameterized function measuring the similarities between sample features and attributes to find the most probable class, and $\Omega$ is used for avoiding overfitting. The generalized $f(\bm{x}_i^s,\bm{W})$ usually takes the following form \cite{DBLP:conf/cvpr/SongSYLS18} \cite{DBLP:conf/nips/FromeCSBDRM13}:
\begin{equation}
\centering
f(\bm{x}_i^s,\bm{W})=\mathop{\arg\max}_{\bm{y}\in\mathcal{Y}}\theta(\bm{x}_i^s)\bm{W}\phi(\bm{y})^{\mathrm{T}}.
\label{e2}
\end{equation}

In fact, $\theta(\cdot)$ refers to the image feature extractor, such as the pre-trained AlexNet \cite{DBLP:conf/nips/KrizhevskySH12}, GoogleNet \cite{DBLP:conf/cvpr/SzegedyLJSRAEVR15} or other well-known deep convolutional network. $\phi(\bm{y})$ is the corresponding attributes of $\bm{y}$.
\subsection{Anchor Generation Model}
In high-dimensional space, it is difficult to distinguish groups belonging to different classes in a irregular manifold shaped like Swiss roll. To produce the discriminative low-dimensional embeddings, some previous unsupervised methods, such as LLE \cite{10.2307/3081722}, utilize the structure information to establish the correlation between samples. In ZSL, the similar samples should be closed to each other in attribute space. Like the user-item matrix in recommendation system, the class-attribute matrix $\bm{C}$ is obviously a bipartite graph where the edges describe the interaction between classes and attributes. To map the classes and attributes into the same embedding space, we merge $\bm{C}$ and $\bm{T}$ into a graph $G=(\mathcal{V},\mathcal{E})$ where $\mathcal{V}$ and $\mathcal{E}$ separately refer to the node set and the edge set. The weighted adjacency matrix is defined as
$\bm{A}^w = 
\left[
\begin{matrix}
\bm{0}_{d_C} & \bm{C} \\
\bm{T}   & \bm{0}_{d_T}
\end{matrix}
\right]\in\mathcal{R}^{(d_T+d_C)\times (d_T+d_C)}$, where $\bm{0}_{d_C}$ denotes the $d_C$ dimensional all-zero matrix. Due to the abundant semantic information, we might as well regard $\bm{F}=\bm{A}^w$ as the node features. In addition, let the diagonal matrix $\bm{D}$ denote degree matrix with $\bm{D}_{ii}=d_i=\sum_j\bm{A}_{ij}^w$ and $\bm{S}=\bm{D}^{-\frac{1}{2}}\bm{A}^w\bm{D}^{-\frac{1}{2}}$ is the normalized adjacency matrix.
To effectively exploit higher-order interactions between classes and attributes, we consider the following diffusion function \cite{DBLP:conf/nips/ZhouBLWS03}:
\begin{equation}
\centering
\mathcal{O}(\bm{H}) = \sum_{i,j}\frac{\bm{A}_{ij}^w}{2}\|\frac{\bm{H}_i}{\sqrt{d_i}}-\frac{\bm{H}_j}{\sqrt{d_j}}\|^2
               + \mu\sum_i\|\bm{H}_i-\bm{F}_i\|^2,
\label{e3}
\end{equation}
where $\bm{H}_i$ denotes the embedding of the $i$-th node. Essentially, the 1-st term implies that the information flows along high weight edges, which forces a node similar to its neighbors. In contrast, the 2-nd term tends to maintain the original features, that is, to preserve the global information. The coefficient $\mu$ constrains the balance between these two terms. Furthermore, the optimal solution is:
\begin{equation}
\centering
\bm{H}^*=(1-\alpha)(\bm{I}-\alpha\bm{S})^{-1}\bm{F},
\label{e4}
\end{equation}
where $\alpha=\frac{1}{1+\mu}$ is a intermediate variable. In spectral graph theory \cite{DBLP:journals/corr/KipfW16}, the convolution operation on graph is formulated as:
\begin{equation}
\centering
\bm{g}_{\theta}*\bm{F}=\bm{V}\bm{G}_{\theta}(\bm{\Lambda})\bm{V}^{\mathrm{T}}\bm{F},
\label{e5}
\end{equation}
where $\bm{g}_{\theta}$ and $\bm{G}_{\theta}(\bm{\Lambda})$ separately denote the spatial filter and the spectral filter, and $\bm{V}$ is the eigenvectors of the normalized Laplacian matrix $\bm{L} = \bm{I} - \bm{S} = \bm{V}\bm{\Lambda} \bm{V}^{\mathrm{T}}$. To avoid to calculating $\bm{V}$, $\bm{G}_{\theta}(\cdot)$ is devised to the polynomial of $\bm{\Lambda}$. In GCN \cite{DBLP:journals/corr/KipfW16}, the reported $\bm{G}_{\theta}(\bm{\Lambda})$ can be finally expressed as:
\begin{equation}
\centering
\begin{aligned}
\bm{G}_{\theta}(\bm{\Lambda}) & =2\bm{I}-\bm{\Lambda}\\
                              & =\bm{I}+(\bm{I}-\bm{\Lambda}),
\end{aligned}
\label{e6}
\end{equation}
where $\bm{\Lambda}_S=\bm{I}-\bm{\Lambda}$ is obviously the eigenvalues of $\bm{S}$. Due to $\bm{\Lambda}_S\in[-1,1]$, it is natural to generalize the Eq. \ref{e6} to a higher-order form:
\begin{equation}
\centering
\begin{aligned}
\bm{G}_{\theta}(\bm{\Lambda}) & =\sum_{k=0}^{\infty}(\alpha\bm{\Lambda}_S)^k\\
                              & =(\bm{I}-\alpha\bm{\Lambda}_S)^{-1}.
\end{aligned}
\label{e7}
\end{equation}

According to the Eq. \ref{e5} and \ref{e7}, we define the diffusion graph convolution following \cite{DBLP:journals/ijon/LiZZCZ19} as:
\begin{equation}
\centering
\begin{aligned}
\bm{g}_{\theta}*\bm{F} & =\bm{V}\bm{G}_{\theta}(\bm{\Lambda})\bm{V}^{\mathrm{T}}\bm{F}\\
                       & = (\bm{I}-\alpha\bm{S})^{-1}\bm{F}.
\end{aligned}
\label{e8}
\end{equation}

Ignoring the scaling coefficient $1-\alpha$, Eq. \ref{e4} has the same form expression as Eq. \ref{e8}. To avoid the overfitting, the truncated form of the Eq. \ref{e8} is employed as follow:
\begin{equation}
\centering
\begin{aligned}
\bm{g}_{\theta}*\bm{F} & =\sum_{k=0}^{p}(\alpha\bm{S})^k\bm{F}.
\end{aligned}
\label{e9}
\end{equation}

As shown in Figure \ref{f3}, the truncated diffusion graph convolution conducts the forward propagation of the auto-encoder to generate the discriminative low-dimensional anchors. This procedure can be described as:
\begin{equation}
\centering
 \bm{U}^{(l+1)}=\sigma(\sum_{k=0}^{p}(\alpha\bm{S})^k\bm{U}^{(l)}\bm{W}^{(l)}),
\label{e10}
\end{equation}
where $\bm{U}^{(l)}$ is the activations of the $l$-th layer and $\bm{W}^{(l)}$ is the weight matrix of the $l$-th layer. In addition, $\sigma(\cdot)$ denotes the activation function. Like the other auto-encoder based models \cite{DBLP:journals/tip/LiaoWL17} \cite{DBLP:conf/kdd/WangC016}, the middle layer activations $\bm{U}=[\bm{U}_C;\bm{U}_T]\in \mathcal{R}^{(d_C+d_T)\times d}$ is extracted as the desired anchors, where $d$ is the number of hidden units.

\subsection{Distribution Alignment Model}
With the learned discriminative low-dimensional anchors, the goal of our model is to project the features $\bm{X}$ to the $\bm{U}$ space while meeting following function:
\begin{equation}
\centering
L_{cons}=\|\bm{X}^s\bm{W}_{cons}-\bm{y}^s\bm{U}_C\|,
\label{e11}
\end{equation}
where $\bm{W}_{cons}$ denotes the matrix projecting the visual features to the $\bm{U}$ space, and $\bm{y}^s\in \mathcal{R}^{N^s\times d_C}$ is the labels corresponding to the $\bm{X}^s$ in the form of one-hot encoding. In fact, $\bm{W}_{cons}$ is a parameterized matrix to measure the similarity between $\bm{X}^s$ and $\bm{y}^s\bm{U}_C$. Like in SAE \cite{DBLP:conf/cvpr/KodirovXG17}, the symmetrical auto-encoder is employed to reconstruct the visual features $\bm{X}^s$ from the hidden states $\bm{\tilde{U}}_C=\bm{X}^s\bm{W}_{cons}$, which can be formulated as:
\begin{equation}
\centering
L_{recons}=\|\bm{\tilde{U}}_C\bm{W}_{recons}-\bm{X}^s\|,
\label{e12}
\end{equation}
where $\bm{W}_{recons}$ is the matrix recovering sample features from the $\bm{U}$ space. Essentially, the Eq. \ref{e12} aligns the distribution between feature space and anchor space through maintaining the information volume. However, as the values in the class-attribute matrix describe the interactions between classes and attributes, the distribution of sample features around each class anchors $\bm{U}_C$ can be regularized by the attribute anchors $\bm{U}_T$. To generate the fine-grained distribution around each anchor, we develop a relation regularization network subjecting to:
\begin{equation}
\centering
L_{reg}=\|\bm{y}^s\bm{C}-\bm{\tilde{U}}_C\bm{M}{\bm{U}_T}^{\mathrm{T}}\|,
\label{e13}
\end{equation}
where $\bm{M}$ is a metric. With described above, we minimize the following loss function:
\begin{equation}
\centering
L=L_{cons}+\lambda_1L_{recons}+\lambda_2L_{reg},
\label{e14}
\end{equation}
where $\lambda_1$ and $\lambda_2$ are weighting coefficients. As shown in Figure \ref{f3}, the distribution alignment component can learn a reliable projection function $\bm{W}_{cons}$. Furthermore, the recognition for unseen classes can be carried out through:
\begin{equation}
\centering
\hat{\bm{y}}=\mathop{\arg\max}_{\bm{y}^*\in \mathcal{L}}\bm{X}^u\bm{W}_{cons}{\bm{y}^*{\bm{U}_C}}^\mathrm{T},
\label{e15}
\end{equation}
where $\mathcal{L}$ is the label set corresponding to the task, i.e., $\mathcal{L}=\mathcal{Y}^u$ in the conventional ZSL setting or $\mathcal{L}=\mathcal{Y}$ in the generalized ZSL setting.

\section{Experiments}
To evaluate the effectiveness of our model, we conduct extensive comparison experiments on several benchmark datasets. Meanwhile, sufficient ablation experiments are carried out to validate the effectiveness of each component.

\subsection{Dataset}
Four benchmark datasets are used in our experiments and the statistics on them is summarized in Table \ref{t1}.

\begin{itemize}
\item[$\bullet$]\textbf{Animals with Attributes 2 (AWA2):} AWA2 \cite{DBLP:journals/corr/XianLSA17} is a coarse-grained dataset containing 37322 images from 50 different classes. In the standard split, 40 source classes are used for training and 10 target classes are used for testing.  For each class, there are about 750 labeled image examples.

\item[$\bullet$]\textbf{Caltech UCSD Birds 200 (CUB):} CUB \cite{WahCUB_200_2011} is a fine-grained and medium scale dataset. It has 11788 images from 200 different types of birds annotated with 312 attributes.

\item[$\bullet$]\textbf{SUN Scene Recognition (SUN):} SUN \cite{DBLP:conf/cvpr/XiaoHEOT10} is also fine-grained, comprising of 717 scenes and 14340 images totally. Commonly, the split with 645 seen classes and 72 unseen classes is used for zero-shot learning.

$\bullet$\textbf{Attribute Pascal and Yahoo (aPY):} aPY \cite{DBLP:conf/cvpr/FarhadiEHF09} is a coarse-grained small-scale dataset with 32 classes and 64 attributes. In our experiment, We follow the split in \cite{DBLP:journals/corr/XianLSA17}, using 20 classes from Pascal and 12 classes from Yahoo as source and target classes respectively.
\end{itemize}

\begin{table}[htbp]
\centering
\caption{Statistics of four tested datasets.}
\begin{tabular}{||c|c|c|c|c||}
\hline
Dataset& \#Attribute& \#Image & Seen/Unseen & \#Dim\\
\hline
\hline
AWA2 & 85 & 37322 & 40/10 & 32\\
CUB  & 312& 11788 & 150/50& 256\\
SUN  & 102& 14340 & 645/72& 64\\
aPY  & 64 & 18627 & 20/12 & 64\\
\hline
\end{tabular}

\label{t1}
\end{table}

\begin{table*}[t]
\centering
\caption{The experimental results on the conventional ZSL. Here the PS and the SS separately refer to the proposed split and the standard split. We report the results (in \%) averaged over 5 repeated experiments. All the methods belong to the inductive ZSL. The best result is marked in \textbf{bold} font. $\mbox{Ours}^{-1}$ and $\mbox{Ours}^{-2}$ refer to the models designed for the ablation experiment.}
\begin{threeparttable}
\begin{tabular}{|p{3cm}<{\centering}|p{1cm}<{\centering}p{1cm}<{\centering}|p{1cm}<{\centering}p{1cm}<{\centering}|p{1cm}<{\centering}p{1cm}<{\centering}|p{1cm}<{\centering}p{1cm}<{\centering}|}
\hline
\hline  
\multirow{2}*{{\bfseries Method}}& \multicolumn{2}{c|}{{\bfseries AWA2}}& \multicolumn{2}{c|}{{\bfseries CUB}}& \multicolumn{2}{c|}{{\bfseries SUN}}&\multicolumn{2}{c|}{{\bfseries aPY}}\\
\cline{2-9}  
& SS & PS& SS & PS& SS & PS& SS & PS\\
\hline
\hline
DAP \cite{DBLP:journals/pami/LampertNH14} & 58.7& 46.1& 37.5& 40.0& 38.9& 39.9& 35.2& 33.8\\
CONSE \cite{DBLP:journals/corr/NorouziMBSSFCD13} & 67.9& 44.5& 36.7& 34.3& 44.2& 38.8& 25.9& 26.9\\
ALE \cite{DBLP:conf/cvpr/AkataPHS13} & 80.3& 62.5& 53.2& 54.9& 59.1& 58.1& -& 39.7\\
ESZSL \cite{DBLP:conf/icml/Romera-ParedesT15} & 75.6& 58.6& 55.1& 53.9& 57.3& 54.5& 34.4& 38.3\\
SJE \cite{DBLP:conf/cvpr/AkataRWLS15}& 69.5& 61.9& 55.3& 53.9& 57.1& 53.7& 32.0& 32.9\\
SYNC \cite{DBLP:conf/cvpr/ChangpinyoCGS16}& 71.2& 46.6& 54.1& 55.6& 59.1& 56.3& 39.7& 23.9\\
SAE \cite{DBLP:conf/cvpr/KodirovXG17}& 80.2& 54.1& 33.4& 33.3& 42.4& 40.3& \textbf{55.4}& 8.3\\
SE-ZSL \cite{DBLP:conf/cvpr/VermaAMR18}& 80.8& 69.2& 60.3& 59.6& 64.5& 63.4& -& -\\
ZSKL \cite{DBLP:conf/cvpr/ZhangK18}& -& 70.5& -& 57.1& -& 61.7& -& \textbf{45.3}\\
F-CLSWGAN \cite{DBLP:conf/cvpr/XianLSA18}& -& -& -& 61.5& -& 62.1& -& -\\
DCN \cite{DBLP:conf/nips/LiuL0J18}& -& -& 55.6& 56.2& \textbf{67.4}& 61.8& -& 43.6\\
PSRZSL \cite{DBLP:conf/cvpr/AnnadaniB18}& -& 63.8& -& 56.0& -& 61.4& -& 38.4\\
\hline
$\mbox{Ours}^{-1}$& 79.3& 67.6& 56.4& 61.5& 53.4& 53.5& 31.7& 23.9\\
$\mbox{Ours}^{-2}$& 76.7& 70.0& 56.5& 61.5& 58.3& 57.9& 32.0& 34.1\\
Ours& \textbf{81.0}& \textbf{73.0}& \textbf{61.4}& \textbf{64.1}& 64.1& \textbf{63.5}& 44.7& 39.2\\
\hline 
\end{tabular}
$\mbox{Ours}^{-1}$: Our model without the graph embedding learning network. \\
$\mbox{Ours}^{-2}$: Our model without the relation metric network.\\
\end{threeparttable}

\label{t2}
\end{table*}

In Table \ref{t1}, \#Dim refers to the dimensions of anchors. For all datasets, their visual features are 2048-dimensional vectors, extracted from the top-layer pooling units of Res101 \cite{DBLP:conf/cvpr/HeZRS16} which is pre-trained on the ImageNet dataset \cite{DBLP:journals/ijcv/RussakovskyDSKS15} without any extra fine-tune. Implementation details can be referred to in \cite{DBLP:journals/corr/XianLSA17}.

\subsection{Settings and Evaluation Metrics}
\textbf{Settings:} Our framework is implemented by using Keras. In the anchor generation stage, we train the graph embedding learning component for 1000 epochs by setting $\alpha=0.8$ and $k=2$. Then, we optimize the distribution alignment component for 1-3 epochs by setting $\lambda_1=1$ and $\lambda_2=5\times 10^{-6}$. These two components are optimized by Adam \cite{DBLP:journals/corr/KingmaB14}. 

For conventional ZSL setting, we evaluate our model on the standard split (SS) as well as on the proposed split (PS) \cite{DBLP:journals/corr/XianLSA17}. It is worth noting that PS is proposed to avoid any test class appearing in ImageNet 1K, i.e., used to train the ResNet model. Thus, the accuracy on PS is crucial to show the generalization of methods. Following PS, we further evaluate DAGDA under the generalized ZSL setting.

\textbf{Evaluation Metrics:} To test the performance of our model, we employ the Mean Class Accuracy (MCA) as our evaluation metric in our experiments:
\begin{equation}
\centering
MCA=\frac{1}{|\mathcal{Y}^t|}\sum_{y\in \mathcal{Y}^t}acc_y,
\label{e16}
\end{equation}
where $|\mathcal{Y}^t|$ is the size of the test label set $\mathcal{Y}^t$ and $acc_y$ is the top-1 accuracy on the data from class $y$. Under the conventional ZSL setting, $\mathcal{Y}^t = \mathcal{Y}^u$ refers to searching only in the unseen label set. However, in the generalized ZSL setting, the search space will be extended to $\mathcal{Y}^t=\mathcal{Y}$, which means that the seen classes $\mathcal{Y}^s$ are included in the test label set as well. Therefore, we evaluate the effectiveness of our model under the generalized ZSL setting by utilizing the harmonic mean (H):
\begin{equation}
\centering
H=\frac{2\times MCA_s\times MCA_u}{MCA_s+MCA_u},
\label{e17}
\end{equation}
where $MCA_s$ denotes the mean class accuracy on the seen test data and $MCA_u$ denotes the mean class accuracy on the unseen test data.

\begin{figure}
\centering
\subfigure[Original attributes]{
\includegraphics[scale=0.6]{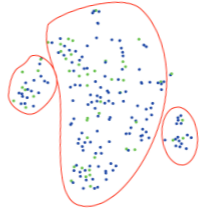}
}
\subfigure[Anchors]{
\includegraphics[scale=0.6]{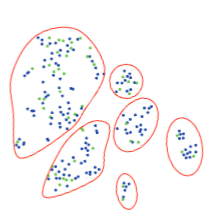}
}
\caption{The t-SNE \cite{maaten2008visualizing} visualization of original attribute space and anchor space on CUB. The blue points and green points separately refer to the source classes and target classes.}
\label{f4}
\end{figure}

\begin{table*}[htbp]
\centering
\caption{The experimental result (in \%) on the generalized ZSL setting. These experiments are conducted with the proposed split. We report results averaged over 5 random trails.}
\begin{tabular}{|c|p{0.8cm}<{\centering}p{0.8cm}<{\centering}p{0.8cm}<{\centering}|p{0.8cm}<{\centering}p{0.8cm}<{\centering}p{0.8cm}<{\centering}|p{0.8cm}<{\centering}p{0.8cm}<{\centering}p{0.8cm}<{\centering}|p{0.8cm}<{\centering}p{0.8cm}<{\centering}p{0.8cm}<{\centering}|}
\hline
\hline  
\multirow{2}*{{\bfseries Method}}& \multicolumn{3}{c|}{{\bfseries AWA2}}& \multicolumn{3}{c|}{{\bfseries CUB}}& \multicolumn{3}{c|}{{\bfseries SUN}}&\multicolumn{3}{c|}{{\bfseries aPY}}\\
\cline{2-13}  
& $MCA_s$ & $MCA_u$& $H$ & $MCA_s$ & $MCA_u$& $H$ & $MCA_s$ & $MCA_u$& $H$ & $MCA_s$ & $MCA_u$& $H$ \\
\hline
\hline
CONSE \cite{DBLP:journals/corr/NorouziMBSSFCD13}& 90.6& 0.5& 1.0& \textbf{72.2}& 1.6& 3.1& 39.9& 6.8& 11.6& \textbf{91.2}& 0.0& 0.0\\
CMT \cite{DBLP:conf/nips/SocherGMN13}& 90.0& 0.5& 1.0& 49.8& 7.2& 12.6& 21.8& 8.1& 11.8& 74.2& 10.9& 19.0\\
SJE \cite{DBLP:conf/cvpr/AkataRWLS15}& 73.9& 8.0& 14.4& 59.2& 23.5& 33.6& 30.5& 14.7& 19.8& 55.7& 3.7& 6.9\\
ESZSL \cite{DBLP:conf/icml/Romera-ParedesT15}& 77.8& 5.9& 11.0& 63.8& 12.6& 21.0& 27.9& 11.0& 15.8& 70.1& 2.4& 4.6\\
SYNC \cite{DBLP:conf/cvpr/ChangpinyoCGS16}& 90.5& 10.0& 18.0& 70.9& 11.5& 19.8& 43.3& 7.9& 13.4& 66.3& 7.4& 13.3\\
SAE \cite{DBLP:conf/cvpr/KodirovXG17}& 82.2& 1.1& 2.2& 54.0& 7.8& 13.6& 18.0& 8.8 & 11.8& 80.9& 0.4& 0.9\\
LATEM \cite{DBLP:conf/cvpr/XianA0N0S16}& 77.3& 11.5& 20.0& 57.3& 15.2& 24.0& 28.8& 14.7& 19.5& 73.0& 0.1& 0.2\\
ALE \cite{DBLP:conf/cvpr/AkataPHS13}& 81.8& 14.0& 23.9& 62.8& 23.7& 34.4& 33.1& 21.8& 26.3& 73.7& 4.6& 8.7\\
ZSKL \cite{DBLP:conf/cvpr/ZhangK18}& 82.7& 18.9& 30.8& 52.8& 21.6& 30.6& 31.4& 20.1& 24.5& 76.2& 10.5& 18.5\\
PSRZSL \cite{DBLP:conf/cvpr/AnnadaniB18}& 73.8& \textbf{20.7}& \textbf{32.3}& 54.3& 24.6& 33.9& 37.2& 20.8& 26.7& 51.4& 13.5& 21.4\\
DCN \cite{DBLP:conf/nips/LiuL0J18}& -& -& -& 37.0& \textbf{25.5}& 30.2& \textbf{60.7}& \textbf{28.4}& \textbf{38.7}& 75.0& 14.2& 23.9\\
\hline
Ours& \textbf{91.5}& 16.5& 28.0& 70.0& 23.5& \textbf{35.2}& 31.0& 14.9& 20.1& 74.1& \textbf{15.5}& \textbf{25.6}\\
\hline 
\end{tabular}

\label{t3}
\end{table*}

\subsection{Conventional Zero-Shot Learning}

For the conventional ZSL setting, we first train the anchor generation component to produce embeddings for each attribute and class. Furthermore, with the standard split and the proposed split, we train the distribution alignment component which projects the source data $\bm{X}^s$ to the source class anchors $\bm{U}_C^s$. We then use the target data $\bm{X}^u$ as the input of the distribution alignment component to produce sample representations in the $\bm{U}$ space. We carry out the classification according to Eq. \ref{e15}. The mean class accuracy (MCA) is employed as the evaluation metric in the conventional ZSL setting.

With the above settings, we evaluate the performance of our model on four benchmark datasets (AWA2, CUB, SUN and aPY), and report the results in Table \ref{t2}. To alleviate the randomness caused by the random initialization, the shown result is the average of 10 experimental results. The experimental results show that our DAGDA outperforms the existing state-of-the-art methods on several datasets (AWA2, CUB and SUN) under the proposed split. In the aPY dataset, our model fails to exceed some methods since only the small-scale graph (32 classes and 64 attributes) can be utilized for graph embedding learning. The goal of our approach is to rectify the distribution of original templates through producing discriminative low-dimensional anchors. In Figure \ref{f4}, the t-SNE visualization from (a) to (b) shows that our model can transform original attributes to more discriminative anchors. Like in the conventional image classification tasks using CNN models, the discriminative features are usually generated in the top-layer pooling units, which is certainly beneficial for the classification. Due to the utilization of the abundant structure information in the class-attribute graph, our model is able to learn the discriminative anchor groups in an unsupervised manner.

\subsection{Ablation Studies}
\begin{figure}
\centering
\subfigure[The architecture of the $\mbox{Ours}^{-1}$.]{
\includegraphics[scale=0.6]{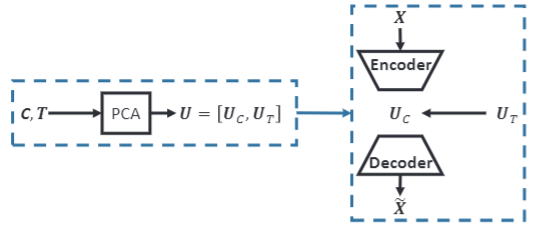}
}
\subfigure[The architecture of the $\mbox{Ours}^{-2}$.]{
\includegraphics[scale=0.6]{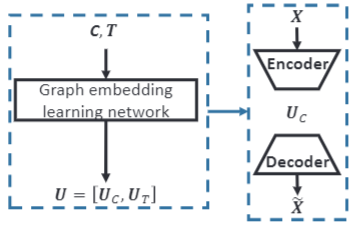}
}
\caption{The models used in the ablation experiments. (a) and (b) are separately corresponding to the $\mbox{Ours}^{-1}$ and the $\mbox{Ours}^{-2}$ in Table \ref{t2}.}
\label{f5}
\end{figure}

To further validate the effectiveness of the anchor generation component and the distribution alignment component, we conduct the ablation experiments shown in Table \ref{t2} where $\mbox{Ours}^{-1}$ refers to our model that replaces the graph embedding learning network with PCA and $\mbox{Ours}^{-2}$ refers to our model that drops the $L_{reg}$ in the  distribution alignment stage. The elaborate architectures of the above two models are shown in Figure \ref{f5}. As expected, the experimental results demonstrate that these two components certainly improve the performance of our model.

\subsection{Generalized Zero-Shot Learning}

In real-world applications, we cannot ensure whether the test samples are from unseen classes. Thus, the generalized ZSL is more convincing to demonstrate the generalization of models than the conventional ZSL. Hence, we evaluate the performance of our model under the generalized ZSL setting. In experiments, we still use the proposed split to divide the data into the source classes and the target classes.

As shown in Table \ref{t3}, the proposed DAGDA outperforms existing state-of-the-art models on the CUB and aPY datasets in terms of $H$. On AWA2, the $MCA_u$ and $H$ are slightly less than the state-of-the-art methods. In the SUN dataset, there are only 20 samples in each class, which is insufficient for the distribution alignment component to learn a fine-grained distribution in the embedding space. Therefore, our model shows a poor performance on SUN. But in general, our model has good generalization capability.

\subsection{Sensitivity Analysis}

\begin{figure}[htbp]
\centering
\subfigure[AWA2]{
\includegraphics[scale=0.6]{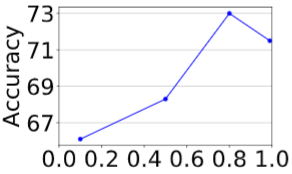}
}
\subfigure[CUB]{
\includegraphics[scale=0.6]{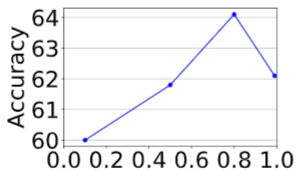}
}
\subfigure[SUN]{
\includegraphics[scale=0.6]{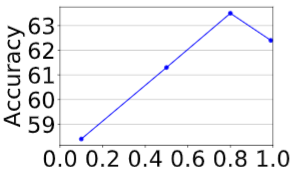}
}
\subfigure[aPY]{
\includegraphics[scale=0.6]{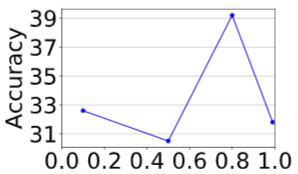}
}
\caption{The model sensitivity about $\alpha$. All the experiments are carried out on the proposed split.}
\label{f6}
\end{figure}

In the Eq. \ref{e10}, $\alpha$ is an important coefficient which can dominate the information aggregation on graph. As shown in Figure \ref{f6}, $\alpha$ certainly affects the performance of our model. If $\alpha$ is closed to $0$, it is difficult to make full use of structural information. In contrast, if $\alpha$ approximates $1$, the local information will be overemphasized to degrade the performance.

%
%
%

In order to obtain as stable a result as possible, we empirically let $\alpha=0.8$ cross all the experiments. The experimental results demonstrate that $\alpha=0.8$ is a good choice.
\section{Further Study and Discussions}
In this paper, we propose the DAGDA model to rectify the irregular distribution of templates in ZSL. More specifically, our model learns a more discriminative anchor space than the original attribute space, while preserving the relations between classes and attributes in sample representation space. In addition, extensive experimental results strongly demonstrate the effectiveness of our model. Meanwhile, it shows that utilizing the interaction between classes and attributes is a nice solution for ZSL.

Essentially, graph based methods aim to model the interactions between entities. In our methods, the classes and attributes are regarded as entities on graph. We develop a diffusion graph convolutional auto-encoder to generate anchors. Due to the effective utilization of the structure information, the generated anchors can be organized according to the graph structure, i.e., some subgraphs will be formed. Therefore, the discriminative anchor space can be produced through above processes. 

In fact, we regard the class embedding as anchor. We train the distribution alignment component so that samples are closed to the corresponding anchors with a reasonable distribution. More specifically, we use attribute anchors to regularize the distribution around each class anchor.

There are still many challenges in zero-shot learning. In further work, we will continue exploiting the graph based zero-shot learning models. We will try to extend our model to implicit side information based rather than attribute based ZSL. In addition, we will research the more effective graph embedding learning method and apply it to zero-shot learning.
\section{Conclusion}
In this paper, we propose a novel framework DAGDA to regularize the distribution of templates in ZSL. The key to solving this problem is to learn a discriminative anchor space. Thus, we develop a graph embedding learning methods to learn the low-dimensional embedding for each class and attribute. With the graph convolutional auto-encoder architect, a truncated diffusion propagation model is raised to conduct the forward propagation. This leads to the generation of the discriminative anchors. To further improve the generalization capability, we develop a distribution alignment component. With the learned anchors, we utilize the interactions between classes and attributes to regularize the distribution of the images in sample embedding space. We carry out extensive experiments to test the effectiveness of our model including the comparative experiments, ablation experiments, and visualization experiments. The experimental results strongly demonstrate that our DAGDA model significantly alleviates aforementioned problem in zero-shot learning.

{\small
\bibliographystyle{IEEEtran}
\bibliography{egbib}
}

\end{document}